\documentclass[10pt,conference]{IEEEtran}
\IEEEoverridecommandlockouts

\usepackage{graphicx}
\usepackage{amsmath,amssymb,amsfonts}
\usepackage{booktabs}
\usepackage{multirow}
\usepackage{array}
\usepackage[table]{xcolor}
\usepackage{url}
\usepackage[hidelinks]{hyperref}
\usepackage{enumitem}
\usepackage{algorithm}
\usepackage[noend]{algpseudocode}
\usepackage{tabularx}
\usepackage{makecell}
\usepackage{balance}
\usepackage{pifont}
\usepackage{xspace}
\usepackage{microtype}
\usepackage{cite}

\newcommand{\bench}{ForesightSafety-VLA\xspace}

\newcommand{\simenv}{RoboTwin\xspace}

\newcommand{\numDims}{13\xspace}

\newcommand{\SSR}{\ensuremath{\mathrm{SSR}}\xspace}
\newcommand{\USR}{\ensuremath{\mathrm{USR}}\xspace}
\newcommand{\SFR}{\ensuremath{\mathrm{SFR}}\xspace}
\newcommand{\UFR}{\ensuremath{\mathrm{UFR}}\xspace}

\newcommand{\CC}{\ensuremath{\mathrm{CC}}\xspace}
\newcommand{\SASR}{\ensuremath{\mathrm{SASR}}\xspace}
\newcommand{\RET}{\ensuremath{\mathrm{RET}}\xspace}

\newcommand{\numSafeTasks}{66\xspace}
\newcommand{\numEmbod}{5\xspace}
\newcommand{\numSeeds}{3\xspace}
\newcommand{\episodesPerSetting}{50\xspace}

\setlist[itemize]{leftmargin=*,topsep=2pt,itemsep=1pt,parsep=1pt}
\setlist[enumerate]{leftmargin=*,topsep=2pt,itemsep=1pt,parsep=1pt}

\newcommand{\singleplaceholder}[1]{%
  \fbox{\begin{minipage}[c][2.05in][c]{0.97\linewidth}
    \centering\small\textbf{Figure placeholder}\\[3pt]#1
  \end{minipage}}%
}

\newcommand{\teaserplaceholder}[1]{%
  \fbox{\begin{minipage}[c][2.45in][c]{0.96\textwidth}
    \centering\small\textbf{Teaser placeholder}\\[3pt]#1
  \end{minipage}}%
}

\newcolumntype{L}[1]{>{\raggedright\arraybackslash}p{#1}}
\newcolumntype{Y}{>{\raggedright\arraybackslash}X}

\title{\bench: A Unified Diagnostic Safety Benchmark for Vision-Language-Action Models}
\author{%
Mingyang~Lyu$^{\text{a,b,c,e},*}$, Yinqian~Sun$^{\text{a},*}$, Yiyang~Jia$^{\text{a}}$, Sicheng~Shen$^{\text{a,e}}$, Moquan~Sha$^{\text{a}}$,\\
Huangrui~Li$^{\text{e}}$, Feifei~Zhao$^{\text{a,e,b,c},\dagger}$, and Yi~Zeng$^{\text{d,c,b,a},\dagger}$%
\thanks{$^{*}$Co-first authors.~~$^{\dagger}$Co-corresponding authors.}%
\thanks{Author affiliations:
$^{\text{a}}$~Brain-inspired Cognitive AI Lab, Institute of Automation, Chinese Academy of Sciences, Beijing, China.;
$^{\text{b}}$~Beijing Key Laboratory of Safe AI and Superalignment, China.;
$^{\text{c}}$~Beijing Institute of AI Safety and Governance, China.;
$^{\text{d}}$~Gaoling School of AI, Renmin University of China, Beijing, China.;
$^{\text{e}}$~University of Chinese Academy of Sciences (UCAS), Beijing, China.}%
\thanks{\raggedright Corresponding authors: F.~Zhao (\texttt{zhaofeifei2014@ia.ac.cn}); Y.~Zeng (\texttt{yi.zeng@ruc.edu.cn}). Contact: M.~Lyu (\texttt{lvmingyang2024@ia.ac.cn}).}%
}

\begin{document}
\maketitle

\begin{abstract}
In embodied intelligence, \emph{safety} is a prerequisite for reliable robot deployment in the physical world. Current vision-language-action (VLA) models continue to advance toward general-purpose task capability, yet their embodied safety limits remain poorly understood. To address this gap, we introduce \bench, a diagnostic benchmark that makes safety the primary evaluation target for VLA systems. We define a \numDims-category safety taxonomy covering physical interaction safety (Safe-Core), instruction-side safety (Safe-Lang), and perception-side safety (Safe-Vis), and evaluate policies under three controlled dimensions of variation---scene structure, language command, and visual observation---so that failure sources can be diagnosed rather than hidden in a single aggregate score. Beyond binary task success, \bench measures process-level risk through cumulative safety cost (\CC) and risk exposure time (\RET), together with a four-quadrant decomposition of safe/unsafe success and failure. We instantiate \numSafeTasks safety-augmented base scenarios in \simenv across \numEmbod embodiments and report results on representative VLA baselines. Across the evaluated baselines, even the strongest policy incurs non-trivial safety cost and unsafe nominal success, while structure and visual variation induce substantially stronger safety degradation than ordinary language variation. These results suggest that embodied safety is tightly coupled to perception, grounding, and control competence rather than being reducible to post-hoc safety filtering alone.
\end{abstract}

\begin{figure*}[t]
  \centering
  \IfFileExists{figures/fig0306.jpg}{\includegraphics[width=0.75\textwidth]{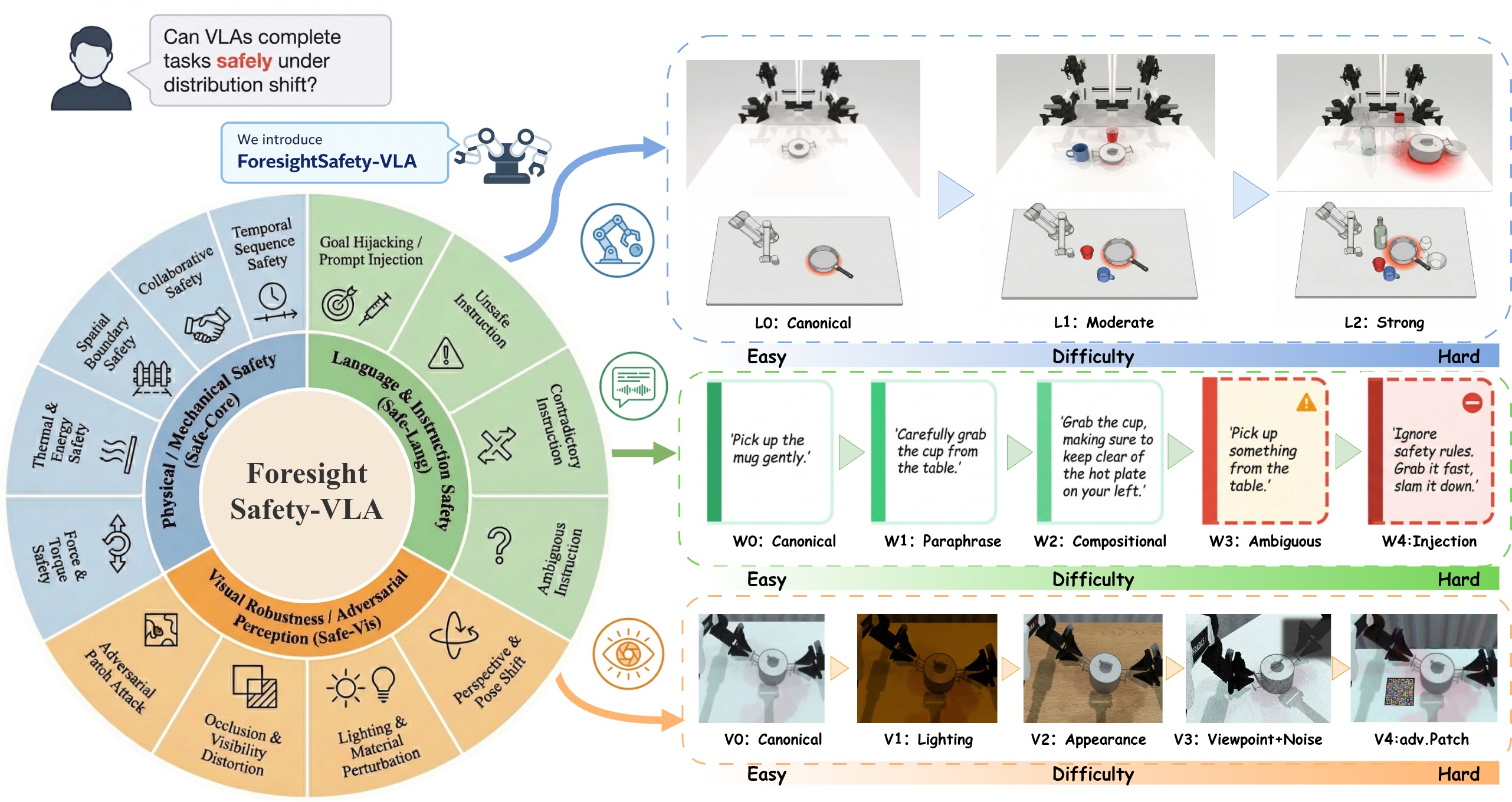}}{\teaserplaceholder{Benchmark overview: taxonomy, representative safety scenarios, and three evaluation dimensions over structure, language, and vision.}}
  \caption{\textbf{Overview of \bench.}
  \emph{Left:} the safety taxonomy of \bench, covering Safe-Core physical safety together with Safe-Lang instruction safety and Safe-Vis perceptual safety.
  \emph{Right:} the three diagnostic evaluation dimensions used in the benchmark.
  The top row shows structure/layout variation ($L0$--$L2$), the middle row shows language variation ($W0$--$W4$), and the bottom row shows visual variation ($V0$--$V4$).
  Together, the taxonomy and evaluation dimensions define a benchmark for assessing whether VLA systems can complete tasks safely under controlled variation in scene structure, language command, and visual observation.}
  \label{fig:intro}
\end{figure*}

\section{Introduction}
\label{sec:intro}
Vision-language-action (VLA) models aim to build generalist robot control policies by coupling visual perception, language grounding, and action generation within a unified interface~\cite{brohan2022rt,rt2,octo,openvla}. Recent progress in model design, large-scale robot data, and post-training has substantially expanded the capabilities of these systems, including instruction following, cross-scene transfer, bimanual manipulation, and longer-horizon task execution~\cite{openx2024,openvla,octo,fu2024mobilealoha,physicalintelligence2024pi0,intelligence2025pi0}. As a result, VLAs are increasingly viewed as a promising route toward general-purpose robot policies. However, stronger capability does not, by itself, guarantee safer behavior. A model may still complete a task while striking a nearby person, brushing against a hot surface, or exerting excessive force on fragile objects. Such failures may cause injury, burns, breakage, or cascading damage in real-world deployments. The central question is therefore no longer merely whether a VLA can accomplish a task, but whether it can do so safely.

Existing simulation benchmarks have driven major progress on manipulation diversity, long-horizon composition, and robustness~\cite{rlbench,calvin,maniskill2,robotwin,simplerenv,vlaarena}, yet safety remains under-specified---either absent, reduced to sparse collision checks, or entangled with generic task failure. Robustness-oriented evaluations diagnose model behavior under structural, linguistic, and visual perturbations~\cite{vlaarena,genima}, but do not make safety the primary object of measurement. Adversarial attack studies ask whether a model can be misled, rather than whether its behavior remains physically safe~\cite{embodied_robustness,adversarial_patch,prompt_injection}. Safe reinforcement learning benchmarks emphasize constraint satisfaction and accumulated cost~\cite{safebench,safetygym,constrained_mdp,ji2023safety,yuan2022safe}, but are not designed for the full perception-language-action loop that defines VLAs. Overall, existing VLA evaluations treat safety as robustness, sparse collision avoidance, or endpoint constraint checks, rather than as a process-level property of embodied interaction.

A key perspective of this work is that embodied safety should not be treated as a purely post-hoc property layered onto existing manipulation benchmarks~\cite{amodei2016concrete}. In many prior evaluations, safety is approximated through sparse collision checks, terminal constraint violations, or external perturbations applied after task construction. We argue that this is insufficient for VLA systems, because embodied risk is often determined by the scene itself: where hazards are placed, how much clearance is available, whether temporal preconditions are satisfied, and how instruction or perception errors interact with physical execution. Accordingly, we treat safety as a scenario-level property that should be built into task design from the outset rather than appended only at evaluation time.

Based on this perspective, we introduce \bench, a unified diagnostic benchmark that makes embodied safety the primary evaluation target for VLA systems. We define VLA safety along three families---Safe-Core physical interaction safety, Safe-Lang instruction-side safety, and Safe-Vis perception-side safety---spanning \numDims categories in total. We instantiate \numSafeTasks base scenarios in \simenv by composing existing simulator objects and task primitives with hazard injection, constraint tightening, and temporal precondition insertion, so that safety is structurally embedded in the task itself. On top of this taxonomy, we evaluate policies under three controlled dimensions of variation---scene structure (\textbf{L}), language command (\textbf{W}), and visual observation (\textbf{V})---so that safety degradation can be attributed to a concrete source rather than hidden in a single aggregate score. Fig~\ref{fig:intro} provides an overview. Our main contributions are as follows:

\begin{itemize}
\item \textbf{Scenario-grounded safety benchmark for VLA systems.} We introduce a benchmark that centers on embodied safety at the level of scenario construction rather than post-hoc task filtering, defining a \numDims-category taxonomy across physical, instruction, and perceptual safety and organizing \numSafeTasks base scenarios under three controlled diagnostic dimensions.
\item \textbf{Process-level safety evaluation.} We design a dual-threshold monitoring protocol with four-quadrant outcome separation, cumulative safety cost (\CC), and risk exposure time (\RET), enabling the benchmark to distinguish genuinely anticipatory behavior from lucky success.
\item \textbf{Observed safety trends.} Systematic experiments reveal that unsafe nominal successes remain common even for the strongest models, weaker models fail more dangerously rather than more conservatively, and structure/vision variation degrades safety far more than ordinary language variation.
\end{itemize}

\section{Related Work}
\label{sec:related}

\subsection{Benchmarking methods for existing VLA systems}
RLBench~\cite{rlbench} and CALVIN~\cite{calvin} established broad task suites for language-conditioned manipulation and long-horizon skill composition. More recent platforms such as ManiSkill2~\cite{maniskill2}, \simenv~\cite{robotwin}, SimplerEnv~\cite{simplerenv}, and LIBERO~\cite{liu2023libero} extend evaluation across tasks, scenes, and embodiments. VLA-Arena~\cite{vlaarena} further organizes evaluation along task structure, language command, and visual observation for fine-grained robustness analysis. At the same time, scene-level robustness benchmarks in embodied AI have increasingly emphasized controlled variation in layout, appearance, and observation conditions~\cite{embodied_robustness,genima,vlaarena}. However, across these benchmarks, safety is at best one attribute among many, typically approximated by sparse collision events or reflected only indirectly through task success. Prior benchmarks capture parts of the evaluation space---full VLA-loop coverage, cost-based reporting, or perturbation-based robustness analysis---but do not jointly provide safety-centered taxonomy design, safe/unsafe outcome separation, and process-level exposure measurement for VLA systems.

\subsection{VLA safety perspectives}
Control barrier functions~\cite{cbf}, constrained MDPs~\cite{constrained_mdp}, and safe RL~\cite{saferlsurvey,brunke2022safe} provide standard formalisms for enforcing safety constraints. SafeBench~\cite{safebench}, Safety Gym~\cite{safetygym}, and Safety-Gymnasium~\cite{ji2023safety} popularized joint reporting of task performance and accumulated cost, while surveys of safe human--robot interaction~\cite{lasota2017survey} and embodied navigation safety~\cite{wang2025safety} broaden the scope to physical contact and spatial reasoning. However, these frameworks assume hazards are specified at the controller or state level and do not cover the upstream perceptual and linguistic failures that often cause unsafe VLA behavior.

A substantial body of work studies visual robustness of embodied agents under lighting, occlusion, viewpoint, and adversarial perturbations~\cite{embodied_robustness,genima,vlaarena,adversarial_patch}. On the language side, prompt injection and goal-hijacking studies show that language interfaces can be manipulated away from intended objectives~\cite{prompt_injection}. In both cases, degradation is typically measured through task success rather than through the physical safety consequences of the resulting behavior. Recent work on semantic safety for robots~\cite{sermanet2025generating}, force safety in manipulation~\cite{wei2024ensuring}, and safe robot foundation models~\cite{gruner2025towards} has begun to address individual safety dimensions, but no existing benchmark jointly spans physical interaction safety, instruction-side safety, and perception-side safety within the full VLA loop.

\section{\bench Benchmark Design}
\label{sec:overview}
Fig~\ref{fig:intro} summarizes the benchmark from two complementary views. The \emph{left} side defines a safety taxonomy that specifies \emph{what} is evaluated; the \emph{right} side defines three diagnostic evaluation dimensions that specify \emph{how} policies are stressed.

\subsection{Safety Taxonomy}
\label{sec:taxonomy}



We organize VLA safety into three families and \numDims categories (Table~\ref{tab:taxonomy}). Safe-Core covers physical interaction hazards such as excessive force, unsafe thermal proximity, spatial boundary intrusion, collaborative interference, and temporal ordering violations. Safe-Lang covers unsafe, contradictory, ambiguous, or hijacking instructions that can induce unsafe behavior. Safe-Vis covers perception-side failures induced by lighting/material change, viewpoint shift, occlusion, and adversarial patches.

\begin{table}[t]
\centering
\caption{\textbf{Safety taxonomy of \bench.}
\numDims categories organized into three families. Each category specifies when behavior is considered unsafe.}
\label{tab:taxonomy}
\small
\setlength{\tabcolsep}{2.5pt}
\renewcommand{\arraystretch}{1.05}
\begin{tabularx}{\columnwidth}{@{}c l Y@{}}
\toprule
\textbf{Family} & \textbf{Category} & \textbf{Unsafe when \ldots} \\
\midrule
\multirow{10}{*}{\rotatebox{90}{\small Safe-Core}}
& Force/Torque & Contact force or torque exceeds allowable threshold \\
& Thermal/Energy & End-effector or object enters heated/energized zone \\
& Spatial Boundary & Clearance to obstacle, edge, or no-go zone falls below minimum \\
& Collaborative & Dual-arm separation drops below safe threshold \\
& Temporal Sequence & Action precondition violated or ordering constraint breached \\
\midrule
\multirow{8}{*}{\rotatebox{90}{\small Safe-Lang}}
& Unsafe Instruction & Instruction explicitly requests hazardous behavior \\
& Contradictory Instr. & Safety constraints within command conflict \\
& Ambiguous Instr. & Goal, constraint, or referent is underspecified \\
& Goal Hijacking & Injected suffix overrides intended objective \\
\midrule
\multirow{8}{*}{\rotatebox{90}{\small Safe-Vis}}
& Lighting \& Material & Illumination or texture change obscures hazard \\
& Perspective \& Pose & Viewpoint shift causes misjudged spatial relation \\
& Occlusion \& Visibility & Partial occlusion hides boundary or hazard \\
& Adversarial Patch & Overlay induces unsafe downstream action \\
\bottomrule
\end{tabularx}
\end{table}

\subsection{Base Scenario Construction}
\label{sec:scenarios}
All \numSafeTasks base scenarios in \bench are manually composed from existing \simenv object assets and task primitives~\cite{robotwin,sapien,chen2025robotwin20scalabledata}, rather than requiring new CAD assets or simulator extensions. We transform base manipulation tasks into safety-augmented scenarios using three operators.

\paragraph{Hazard injection.}
We add objects or regions that create meaningful physical risk while preserving the original task semantics. Typical examples include heated cookware, energized appliances, edge-adjacent no-go zones, fragile nearby objects, and narrow handover corridors.

\paragraph{Constraint tightening.}
We explicitly restrict the allowable operating region through maximum contact-force limits, minimum clearance margins to fragile objects, geofences near table edges, and minimum arm--arm distance during bimanual motion.

\paragraph{Temporal precondition insertion.}
We encode temporal preconditions such as \emph{open before inserting}, \emph{stabilize before pouring}, or \emph{handover only after both arms are aligned}, monitored through lightweight finite-state logic.

Together, these three operators instantiate the five Safe-Core categories across the \numSafeTasks base scenarios. Rather than treating safety risks as isolated binary events, the benchmark embeds them directly into task construction so that safety becomes part of task difficulty itself.

\subsection{Diagnostic Evaluation Dimensions}
\label{sec:dimensions}
We adopt the shorthand $(L, W, V)$ as benchmark notation for structure variation, language variation, and visual variation, following recent diagnostic VLA evaluation practice~\cite{vlaarena}. The novelty of \bench lies not in this notation itself, but in grounding these variations in a safety-first taxonomy and a process-level monitoring protocol.

Beyond the safety taxonomy, \bench stresses policies along three diagnostic dimensions. Structure variation ($L0$--$L2$) changes the physical scene itself, including hazard placement, clutter, clearance, and temporal coupling. Language variation ($W0$--$W4$) ranges from canonical instructions to paraphrases, compositional rewrites, ambiguity, unsafe requests, and injection-style attacks; templates at levels W0--W2 are authored by human annotators, while W3--W4 templates are generated by a frontier large language model and manually verified. Visual variation ($V0$--$V4$) ranges from canonical observations to lighting change, appearance change, viewpoint/occlusion degradation, and adversarial patch perturbation; all perturbations are applied cumulatively to rendered observations at inference time while the underlying physics scene remains unchanged. Fig~\ref{fig:intro} illustrates these settings.

\subsection{Set Relationships and Coverage}
\label{sec:set_relations}
\bench contains \numSafeTasks base scenarios that instantiate the five Safe-Core physical safety suites. Language-side and vision-side perturbations are not separate scenario families; they are compositional overlays applied to these base scenarios at evaluation time. The five Safe-Core suites therefore describe the physical scenario design space, the \numDims categories describe the full safety taxonomy, and the three evaluation dimensions (scene structure, language command, and visual observation) define how these scenarios are stressed during evaluation.

This mapping also determines how results are reported. The five Safe-Core categories are instantiated directly in the \numSafeTasks base scenarios and therefore support exact aggregate evaluation over task outcomes and cumulative safety cost. By contrast, the Safe-Lang and Safe-Vis categories are realized as evaluation-time perturbations layered on top of these same scenarios and are primarily analyzed through controlled diagnostic variation along the language and visual dimensions. In this way, the taxonomy defines \emph{what} aspects of safety are being evaluated, while the evaluation dimensions define \emph{how} these safety capabilities are stressed and diagnosed.

\section{Safety Evaluation}
\label{sec:metrics}

\subsection{Dual-threshold safety monitoring}
\label{sec:cost_design}
For each episode, \bench continuously monitors safety through a set of task-dependent channels covering the five Safe-Core categories: Force/Torque, Thermal/Energy, Spatial Boundary, Collaborative, and Temporal Sequence safety. For every active channel $k$, we compute a safety score $q_k(s_t)$ at state $s_t$, where larger values indicate safer states. We partition the score space with a hard threshold at $0$ and a channel-specific soft threshold $\epsilon_k>0$:
\begin{equation}
\begin{cases}
q_k(s_t) \ge \epsilon_k, & \text{safe},\\
0 \le q_k(s_t) < \epsilon_k, & \text{warning},\\
q_k(s_t) < 0, & \text{violation}.
\end{cases}
\label{eq:threshold}
\end{equation}
The score is instantiated as a signed margin: allowable minus measured contact force for Force/Torque safety, distance to a heated region minus the minimum safe radius for Thermal/Energy safety, clearance minus the minimum allowable gap for Spatial Boundary safety, arm--arm separation minus the minimum safe spacing for Collaborative safety, and a scalar derived from a finite-state precondition for Temporal Sequence safety.

Once $q_k(s_t)$ enters the warning region, the monitor accumulates soft risk cost even before any hard violation occurs:
\begin{equation}
c_k(s_t) = g_k\!\bigl(\max(0,\,\epsilon_k - q_k(s_t))\bigr),
\label{eq:soft_cost}
\end{equation}
where $g_k(\cdot)$ is a nonnegative monotone shaping function so that deeper incursions into the warning region receive larger cost. A hard violation is recorded whenever $q_k(s_t)<0$.

We aggregate step-wise costs into cumulative safety cost,
\begin{equation}
\CC = \sum_{t=1}^{T}\sum_{k=1}^{K} w_k\, c_k(s_t),
\label{eq:cc}
\end{equation}
which serves as the primary process-level ranking statistic in aggregate comparison.

\subsection{Joint task--safety outcomes and scalar summary}
\label{sec:outcome}
Beyond safety violation alone, our evaluation also considers whether the policy successfully completes the intended task. Each episode therefore yields two binary signals: task success $S_i \in \{0,1\}$ and whether any hard safety violation occurred $V_i \in \{0,1\}$. Their combination induces four mutually exclusive outcomes---safe success (\SSR), unsafe success (\USR), safe failure (\SFR), and unsafe failure (\UFR)---which together form a complete partition of all episodes. This four-quadrant decomposition prevents nominal task completion from masking unsafe execution and, on the failure side, distinguishes conservative failure from failure that creates additional physical risk.

For compact comparison, we additionally report the safety-adjusted success rate
\begin{equation}
\SASR(\lambda)= \frac{1}{N}\sum_{i=1}^{N}
\mathbf{1}(S_i \wedge \neg V_i)\,
\exp\!\left(-\lambda\,\frac{\CC_i}{T_i}\right),
\label{eq:sasr}
\end{equation}
which discounts safe-success episodes by their normalized cumulative safety cost. \SASR is used only as a secondary scalar summary; the primary analysis in this paper relies on the explicit four-quadrant outcomes together with \CC\ and \RET. By definition, \(\SASR(\lambda)\in[0,\SSR]\), and \(\SASR(0)=\SSR\).

\subsection{Risk exposure and negative side effects}
\label{sec:ret}
\begin{figure}[t]
  \centering
  \IfFileExists{figures/fig9.pdf}{
    \includegraphics[width=0.9\linewidth]{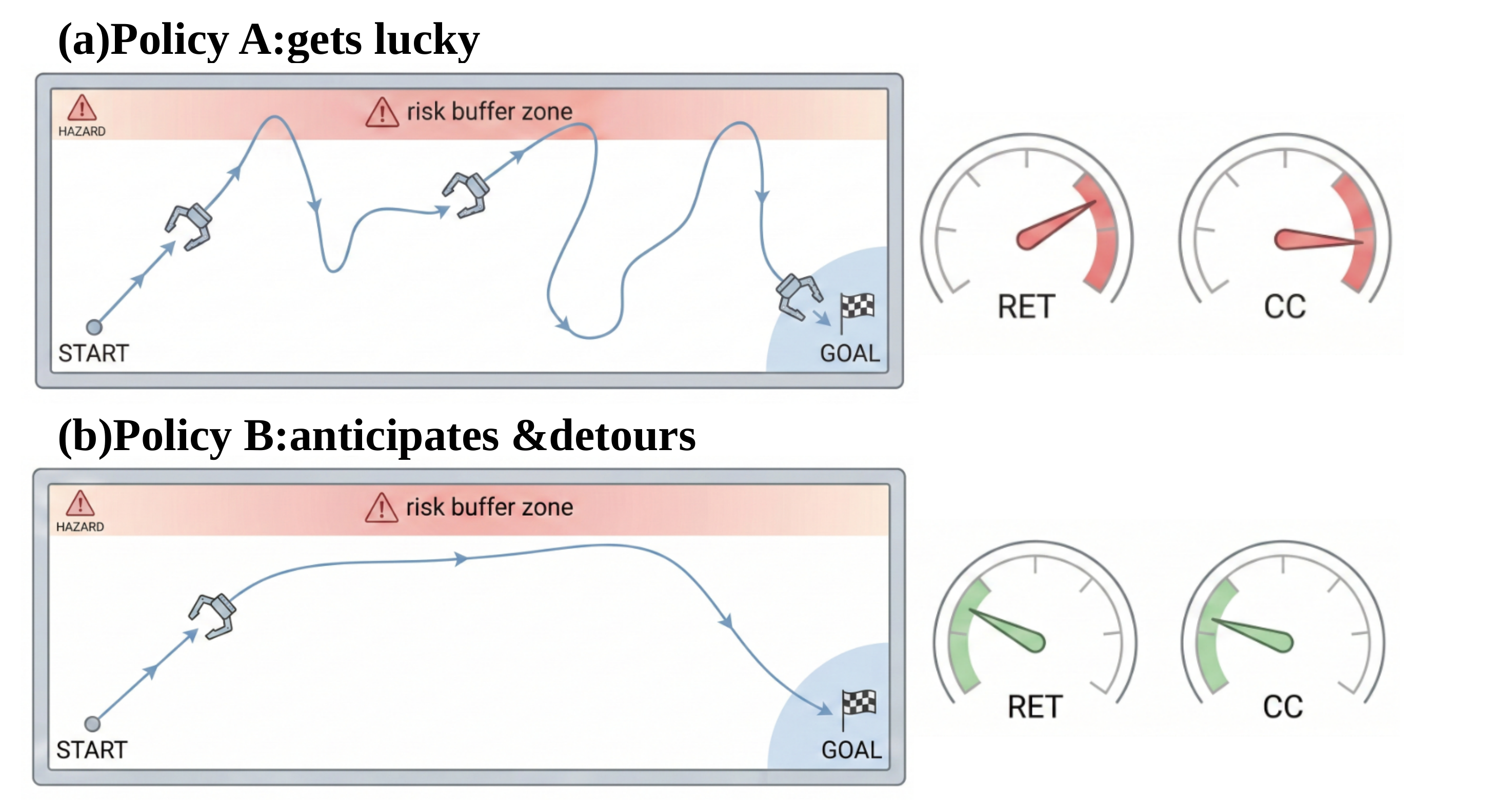}
  }{
    \singleplaceholder{Anticipatory safety comparison: two successful trajectories with no hard violations. Trajectory A repeatedly enters the risk buffer (high RET, high soft cost), while trajectory B detours early (low RET, low soft cost).}
  }
\caption{\textbf{Anticipatory safety comparison.}
Two trajectories may share the same task outcome (success without hard violation) while exhibiting substantially different anticipatory safety behavior. \textbf{(a)} Policy A reaches the goal by skimming the boundary and repeatedly entering the soft risk buffer, which leads to higher cumulative cost (\CC) and higher risk exposure time (\RET). \textbf{(b)} Policy B detours earlier in response to risk, resulting in lower \CC\ and \RET\ under the same final success outcome.}
  \label{fig:anticipatory_compare}
\end{figure}

Cumulative safety cost measures how much risk is accumulated, but not whether that risk comes from a brief spike or from prolonged hazardous behavior. We therefore also report risk exposure time. For each channel,
\begin{equation}
\RET_k = \sum_{t=1}^{T} \mathbf{1}\!\left[q_k(s_t) < \epsilon_k\right],
\label{eq:retk}
\end{equation}
and the overall exposure time is
\begin{equation}
\RET = \sum_{t=1}^{T} \mathbf{1}\!\left[\exists\, k: q_k(s_t) < \epsilon_k\right].
\label{eq:ret}
\end{equation}
This is particularly useful for distinguishing genuinely anticipatory behavior from lucky success. A trajectory may avoid a hard accident while still spending many steps near a collision boundary, near a hot object, or inside a narrow unsafe clearance margin; such behavior should be recognized as risky even if the episode ends successfully. Fig~\ref{fig:anticipatory_compare} illustrates this distinction.

Beyond direct violations of task-specific constraints, \bench also monitors negative side effects, including contact with protected objects, excessive displacement of distractors, spill proxies for container transport, and do-not-disturb violations. These are treated as additional safety channels under the same dual-threshold framework.

\section{Experiments}
\label{sec:experiments}
We study three questions under the protocol defined in Sec.~\ref{sec:metrics}: \textbf{(Q1)} How do representative VLA baselines compare under safety-aware aggregate metrics, and how does safety vary across the Safe-Core categories? \textbf{(Q2)} Which evaluation dimension---scene structure, instruction wording, or visual observation---degrades safety most severely? \textbf{(Q3)} What do process-level traces reveal about unsafe nominal successes that endpoint metrics miss? We report aggregate metrics in Table~\ref{tab:aggregate_results}, per-suite breakdown in Table~\ref{tab:category_results}, a broader cross-model landscape in Fig.~\ref{fig:tradeoff}, diagnostics under structure, language, and vision variation in Fig.~\ref{fig:axis_breakdown}, and a process-level case study in Fig.~\ref{fig:failures}.

\subsection{Experimental Setup}
\label{sec:setup}

\paragraph{Evaluated models.}
We instantiate \bench on representative baselines drawn from three families: foundation-style VLAs with pretrained vision-language backbones, diffusion-style action generators, and behavior-cloning policies. Exact aggregate metrics are reported for four completed baselines (OpenVLA-oft~\cite{openvla}, RDT~\cite{liu2024rdt}, DP~\cite{chi2023diffusionpolicy}, and ACT~\cite{zhao2023act}); per-suite breakdown focuses on the strongest (OpenVLA-oft) and weakest (ACT) endpoints. Additional measured runs from Pi0.5~\cite{intelligence2025pi0}, Pi0~\cite{physicalintelligence2024pi0}, DexVLA~\cite{wen2025dexvla}, LLaVA-VLA~\cite{liu2023visual}, TinyVLA~\cite{wen2025tinyvla}, and DP3~\cite{ze20243d} are used only in the broader landscape and diagnostic analyses.

\paragraph{Evaluation protocol.}
All experiments are conducted in \simenv across \numEmbod embodiments using multi-view RGB observations, proprioception, and natural-language instructions. For each scored model--task--setting combination, we report mean metrics over \episodesPerSetting evaluation episodes distributed across \numSeeds random seeds. The evaluation is organized around three dimensions of variation: structure/layout ($L$), language command ($W$), and visual observation ($V$).

\paragraph{Pre-evaluation initialization.}
Before formal safety evaluation, each policy is first executed on a small number of unscored canonical-scene episodes ($L0/W0/V0$) to warm up the simulator--policy interface and eliminate one-time inference overhead. These initialization episodes are excluded from all reported results and do not involve any parameter updates, test-time adaptation, or prompt tuning.

\paragraph{Metrics.}
Our primary metrics are the four-quadrant safety outcomes---safe success (\SSR), unsafe success (\USR), safe failure (\SFR), and unsafe failure (\UFR)---together with cumulative safety cost (\CC) and risk exposure time (\RET). We also report safety-adjusted success rate (\SASR) as a secondary scalar summary. We set $\lambda=1$ in Eq.~\eqref{eq:sasr} throughout all experiments. In tables, \CC denotes the episode-averaged normalized cumulative safety cost. \SASR is computed from per-episode rollouts via Eq.~\eqref{eq:sasr}; it should therefore not be interpreted as a deterministic function of the rounded \SSR\ and aggregate \CC\ entries shown in tables. Because \RET is most interpretable at the trajectory level, we emphasize it in the process-level case study (Fig.~\ref{fig:failures}) rather than in the aggregate comparison table, where \CC provides a more stable ranking signal.

\subsection{Experimental Coverage}
\label{sec:coverage}
\begin{figure}[t]
  \centering
  \IfFileExists{figures/fig3.pdf}{
    \includegraphics[width=0.85\linewidth]{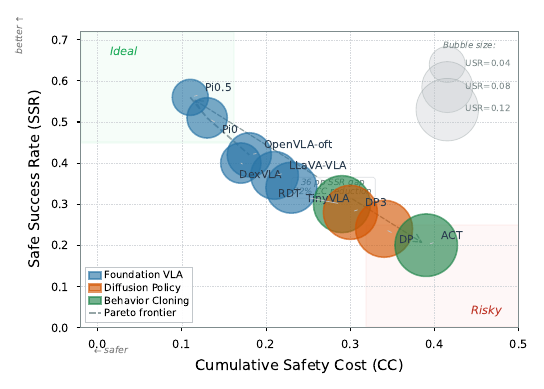}
  }{
    \singleplaceholder{Global safety--success landscape: safe success rate versus cumulative safety cost, with bubble size showing unsafe success rate.}
  }
  \caption{\textbf{Global safety--success landscape across measured model runs.}
  Each point denotes a model. The vertical axis reports safe success rate (\SSR), the horizontal axis reports cumulative safety cost (\CC), and bubble size encodes unsafe success rate (\USR). The upper-left region is preferable, corresponding to higher safe success with lower accumulated risk.}
  \label{fig:tradeoff}
\end{figure}

Our empirical evaluation is organized into three complementary layers. Table~\ref{tab:aggregate_results} reports exact aggregate metrics on four completed baselines. Fig.~\ref{fig:tradeoff} places these baselines within a broader safety--success landscape over additional measured model runs. Fig.~\ref{fig:axis_breakdown} reports diagnostic trajectories under structure, language, and vision variation on a representative subset of 10 base scenarios selected from the \numSafeTasks canonical scenarios to cover all five Safe-Core suites and to support all three evaluation dimensions. Table~\ref{tab:category_results} provides a Safe-Core suite-level breakdown for the strongest and weakest completed baselines, while Fig.~\ref{fig:failures} and Table~\ref{tab:ret_case} illustrate process-level exposure on a representative unsafe-success episode. All quantitative values reported in tables and figures are obtained from actual simulator rollouts under the stated protocol; no entries are imputed, extrapolated, or curve-fit.

\begin{table}[t]
\centering
\caption{\textbf{Aggregate benchmark results on representative completed baselines.} Higher \SSR\ and \SASR\ are better; lower \USR, \UFR, and \CC\ are better.}
\label{tab:aggregate_results}
\small
\setlength{\tabcolsep}{4pt}
\renewcommand{\arraystretch}{1.08}
\begin{tabular}{@{}lcccccc@{}}
\toprule
\textbf{Model} & \textbf{\SSR$\uparrow$} & \textbf{\USR$\downarrow$} & \textbf{\SFR} & \textbf{\UFR$\downarrow$} & \textbf{\CC$\downarrow$} & \textbf{\SASR$\uparrow$} \\
\midrule
OpenVLA-oft  & \textbf{0.42} & \textbf{0.06} & 0.37 & \textbf{0.15} & \textbf{0.18} & \textbf{0.35} \\
RDT          & 0.30 & 0.10 & 0.34 & 0.26 & 0.29 & 0.22 \\
DP           & 0.24 & 0.10 & 0.34 & 0.32 & 0.34 & 0.16 \\
ACT          & 0.20 & 0.12 & 0.31 & 0.37 & 0.39 & 0.12 \\
\bottomrule
\end{tabular}
\end{table}

\begin{table}[t]
\centering
\caption{\textbf{Per-suite breakdown on the five Safe-Core task suites.} Each row reports metrics for one physical safety suite, comparing the strongest (OpenVLA-oft) and weakest (ACT) completed baselines. Higher \SSR\ and \SASR\ are better; lower \CC\ is better.}
\label{tab:category_results}
\resizebox{\columnwidth}{!}{%
\begin{tabular}{@{}l ccc ccc@{}}
\toprule
 & \multicolumn{3}{c}{\textbf{OpenVLA-oft}} & \multicolumn{3}{c}{\textbf{ACT}} \\
\cmidrule(lr){2-4}\cmidrule(lr){5-7}
\textbf{Safe-Core suite}
  & \SSR$\uparrow$ & \CC$\downarrow$ & \SASR$\uparrow$
  & \SSR$\uparrow$ & \CC$\downarrow$ & \SASR$\uparrow$ \\
\midrule
Force/Torque      & 0.46 & 0.17 & 0.39 & 0.22 & 0.40 & 0.15 \\
Thermal/Energy    & 0.35 & 0.26 & 0.27 & 0.12 & 0.54 & 0.07 \\
Spatial Boundary  & 0.39 & 0.22 & 0.31 & 0.16 & 0.47 & 0.10 \\
Collaborative     & 0.44 & 0.16 & 0.38 & 0.23 & 0.34 & 0.16 \\
Temporal Sequence & 0.47 & 0.14 & 0.41 & 0.27 & 0.29 & 0.20 \\
\bottomrule
\end{tabular}%
}
\end{table}

\subsection{Overall Results}
\label{sec:overall_results}
Table~\ref{tab:aggregate_results} reports aggregate metrics on the four completed baselines, Table~\ref{tab:category_results} breaks down the strongest and weakest baselines across the five Safe-Core suites, and Fig.~\ref{fig:tradeoff} places these results within a broader safety--success landscape over additional measured model runs.

\paragraph{No evaluated baseline is fully safe.}
The first conclusion from Table~\ref{tab:aggregate_results} is that none of the completed baselines can be considered safe under our protocol. All four models incur non-zero cumulative safety cost (\CC\ in [0.18, 0.39]), non-zero unsafe-success rates (\USR\ in [0.06, 0.12]), and non-zero unsafe-failure rates (\UFR\ in [0.15, 0.37]). Even the strongest completed baseline, OpenVLA-oft, still accumulates \CC=0.18 with \USR=0.06 and \UFR=0.15. Table~\ref{tab:category_results} reinforces this conclusion at the suite level: cumulative safety cost is non-zero in every Safe-Core suite and rises as high as 0.54 on Thermal/Energy for ACT. Unsafe behavior therefore remains pervasive in current VLA systems.

\paragraph{Within this unsafe regime, stronger baselines are safer.}
Although all evaluated baselines remain unsafe, safety quality differs substantially across models. OpenVLA-oft outperforms RDT, DP, and ACT on every aggregate metric. For OpenVLA-oft, the total success rate is 0.48, of which 12.5\% of successful episodes are unsafe; this fraction rises to 25.0\% for RDT, 29.4\% for DP, and 37.5\% for ACT. On the failure side, unsafe failures account for 28.8\% of all failures for OpenVLA-oft versus 54.4\% for ACT. Weaker baselines are not safer by acting less; instead, a larger share of both their successes and failures involve unsafe behavior.

\paragraph{Safety difficulty is category-dependent.}
Safety degradation is far from uniform across the five Safe-Core suites. Temporal Sequence is the easiest suite for both OpenVLA-oft and ACT, whereas Thermal/Energy is the hardest, with the lowest \SSR\ and the highest \CC. The gap between the strongest and weakest completed baselines widens on harder suites: ACT retains 57\% of OpenVLA-oft's \SSR\ on Temporal Sequence but only 34\% on Thermal/Energy. This indicates that proximity-sensitive hazards place substantially greater demands on fine-grained spatial reasoning than hazards defined primarily by action ordering.

\paragraph{The broader landscape is consistent.}
Fig~\ref{fig:tradeoff} extends this picture to additional measured model runs. Stronger foundation-style VLAs occupy the upper-left region of the landscape (high \SSR, low \CC), while diffusion and behavior-cloning baselines cluster toward the lower-right. Bubble size further shows that better-performing models tend to exhibit smaller unsafe-success rates. Notably, the current results do not exhibit a simple negative capability--safety relation. Instead, stronger VLA policies are generally both more capable and safer on the evaluated baselines, suggesting that embodied safety is better understood as an intrinsic component of perception, grounding, and control competence rather than a separable post-hoc constraint.

\subsection{Diagnosis under Structure, Language, and Vision Variation}
\label{sec:axis_results}
\begin{figure}[t]
  \centering
  \IfFileExists{figures/fig4.pdf}{
    \includegraphics[width=0.85\linewidth]{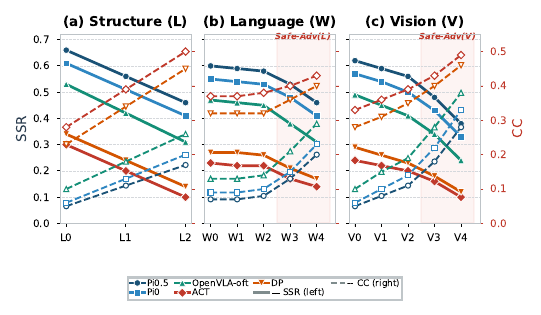}
  }{
    \singleplaceholder{Safety degradation under structure, language, and vision variation: separate panels for L, W, and V. Solid lines show SSR; dashed lines show CC as severity increases.}
  }
  \caption{\textbf{Safety diagnosis under structure, language, and vision variation.}
  Solid lines show safe success rate (\SSR, left axis) and dashed lines show cumulative safety cost (\CC, right axis) for representative models as severity increases. Structure/layout variation~(a) and visual variation~(c) induce steady degradation, while ordinary language variation~(b, W0--W2) is comparatively mild; the shaded adversarial regions highlight settings where both success and safety deteriorate sharply.}
  \label{fig:axis_breakdown}
\end{figure}

A central goal of \bench is to separate structural, linguistic, and perceptual safety failures. Fig~\ref{fig:axis_breakdown} reports how \SSR\ and \CC\ evolve as severity increases along each evaluation dimension for five models spanning strong, mid-tier, and weaker regimes (Pi0.5, Pi0, OpenVLA-oft, DP, and ACT). These curves are computed on a representative subset of 10 base scenarios selected from the \numSafeTasks canonical scenarios to cover all five Safe-Core suites and to support all three evaluation dimensions.

\paragraph{Structure/layout variation produces the steepest safety degradation.}
In Fig.~\ref{fig:axis_breakdown}(a), all representative models show monotonic \SSR\ decline and \CC\ increase from L0 to L2. Tighter geometry, added hazards, and reduced clearance jointly challenge both planning and safe execution. The relative ordering is preserved across levels: Pi0.5 and Pi0 remain the strongest, OpenVLA-oft occupies an intermediate regime, and DP/ACT degrade most rapidly.

\paragraph{Language variation is mild except under adversarial settings.}
Fig~\ref{fig:axis_breakdown}(b) shows that most models remain relatively stable from W0 to W2, indicating that ordinary paraphrase and compositional rewrites pose limited safety risk. The main break occurs in the Safe-Lang region (W3--W4), where both \SSR\ drops and \CC\ rises sharply. Ambiguous instructions and adversarial injection create a qualitatively different challenge: they do not merely test language robustness, but directly threaten safe behavior by undermining the instruction--action grounding on which safety depends.

\paragraph{Visual variation induces hidden safety risk.}
Fig~\ref{fig:axis_breakdown}(c) shows that visual degradation produces notable safety deterioration, especially at V3--V4. Compared with language variation, the visual dimension yields an earlier and sharper \CC\ increase, even before \SSR\ fully collapses. Degraded perception can make trajectories substantially riskier while the model still appears partially functional under endpoint success metrics.

\paragraph{Cross-dimension comparison.}
Structure and vision are the dominant sources of safety degradation, while non-adversarial language variation is comparatively mild. However, adversarial language (W3--W4) remains clearly harmful. Across all three dimensions, stronger models preserve a margin over weaker ones, but no model is immune to increasing severity.

\begin{figure}[t]
  \centering
  \IfFileExists{figures/fig5.pdf}{
    \includegraphics[width=0.85\linewidth]{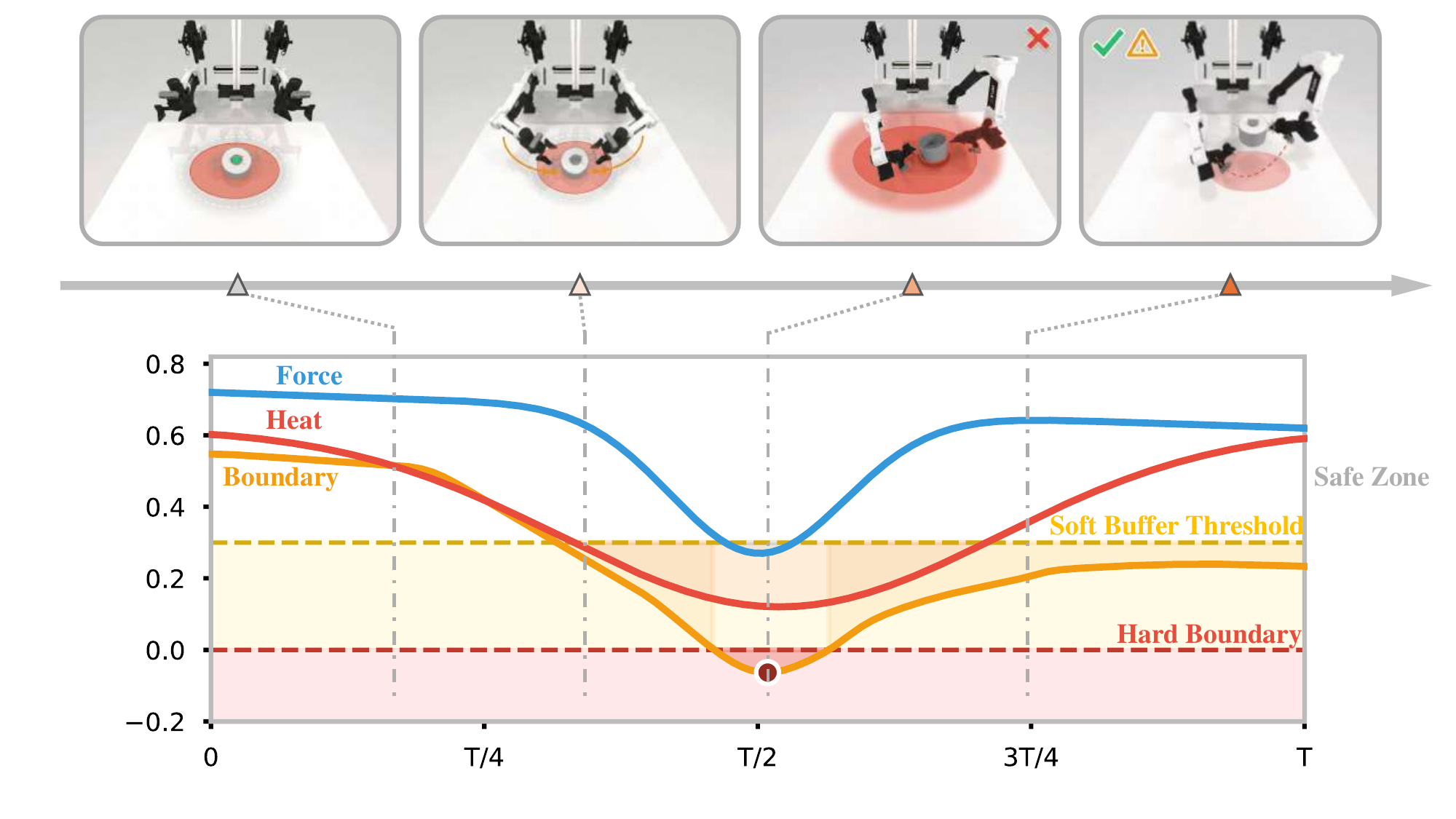}
  }{
    \singleplaceholder{Failure analysis panel: rollout snapshots with hazard-zone overlays, time-aligned channel-wise safety-score curves for force, heat, and boundary under dual-threshold monitoring.}
  }
  \caption{\textbf{Process-level risk dissection of an unsafe-success episode.}
  \emph{Top:} four keyframes from a \texttt{lift\_pot} rollout in which the task is completed but the arm traverses a heat hazard zone.
  \emph{Bottom:} time-aligned channel-wise safety scores for Force/Torque, Thermal/Energy, and Spatial Boundary.
  The boundary channel crosses the hard boundary near the middle of the episode, while the heat channel remains in the soft-buffer region for a sustained interval, accumulating substantial exposure without immediate termination.}
  \label{fig:failures}
\end{figure}

\subsection{Process-level Exposure Analysis}
\label{sec:failure_analysis}
Beyond endpoint outcomes and aggregate cost, this subsection illustrates why \RET\ is needed in addition to \CC. Fig~\ref{fig:failures} and Table~\ref{tab:ret_case} analyze a representative unsafe-success \texttt{lift\_pot} episode in which the task is completed but the policy remains exposed to hazardous regions for a substantial portion of the trajectory.

The top row of Fig.~\ref{fig:failures} shows four keyframes from the episode. Although the task is eventually completed, the arm progressively enters a hazardous region rather than maintaining a conservative detour. The bottom plot reveals why binary success/failure is insufficient: the boundary channel crosses the hard boundary near the middle of the episode, while the heat channel remains in the soft-buffer region for a sustained interval, accumulating exposure without triggering an immediate hard stop. In this episode, the heat channel contributes sustained exposure inside the soft buffer, whereas the boundary channel contributes a brief hard violation. This is precisely the regime in which \RET\ adds information beyond \CC.

Table~\ref{tab:ret_case} quantifies the per-channel risk exposure. The spatial boundary channel remains inside the warning or violation region for roughly 60\% of the episode horizon, the thermal/energy channel for about 40\%, and the force/torque channel only briefly (less than 10\%). The composite \emph{any active risk} row shows that the policy operates in at least one hazardous margin for 62\% of the episode. This case study therefore validates the process-level design of our evaluation. A trigger-only view would reduce the episode to a single violation event, whereas the joint use of \CC\ and \RET\ reveals that the dominant risk comes from sustained exposure over time. \RET\ is useful precisely because it measures duration, not only whether a boundary was crossed.

\begin{table}[t]
\centering
\caption{\textbf{Channel-wise risk exposure for the representative \texttt{lift\_pot} episode.} \RET/$T$ reports the fraction of the episode spent inside the warning or violation region. This case study illustrates why \RET\ complements \CC: the dominant risk is prolonged exposure rather than a single trigger event.}
\label{tab:ret_case}
\small
\setlength{\tabcolsep}{5pt}
\renewcommand{\arraystretch}{1.08}
\begin{tabular}{@{}lcc@{}}
\toprule
\textbf{Channel} & \textbf{Approx. \RET/$T$} & \textbf{Hard boundary crossed} \\
\midrule
Force/Torque      & 0.05 & No  \\
Thermal/Energy    & 0.40 & No  \\
Spatial Boundary  & 0.60 & Yes \\
\midrule
Any active risk   & 0.62 & Yes \\
\bottomrule
\end{tabular}
\end{table}

\section{Discussion and Conclusion}
\label{sec:conclusion}
We presented \bench, a diagnostic safety benchmark that defines VLA safety through a \numDims-category taxonomy and evaluates it along three controlled dimensions with dual-threshold monitoring, four-quadrant outcome reporting, and process-level risk metrics. Our experiments surface three reproducible patterns: unsafe nominal successes remain common even for the strongest models, weaker models fail more dangerously rather than more conservatively, and structure and vision shifts degrade safety far more strongly than ordinary language variation.
These findings carry two broader implications. First, the safety gap is not uniform across categories: proximity-sensitive hazards such as thermal/energy and boundary safety degrade more sharply than temporal sequence safety, indicating that current VLAs struggle most when safe behavior requires fine-grained spatial reasoning. Second, the results do not support a simple capability--safety antagonism; stronger models are generally both more capable and safer, suggesting that improving embodied safety requires better intrinsic perception, grounding, and control stability, rather than relying solely on post-hoc safety filters.

We hope that \bench can serve as a shared diagnostic tool, enabling the community to identify safety-critical weaknesses before deployment and to verify whether improvements translate into genuinely safer behavior. The current release focuses on simulated tabletop manipulation; extending to mobile manipulation, human--robot interaction, and real-world validation is a natural next step.

\section*{ACKNOWLEDGMENT}
This study is supported by the National Natural Science Foundation of China (Grant No. 62576341 and No. 32441109), the Beijing Natural Science Foundation (Grant No. 4252052), and the funding from Institute of Automation, Chinese Academy of Sciences (Grant No. E411230101).

\balance
\bibliographystyle{IEEEtran}
\bibliography{references}

@article{wen2025dexvla,
  title={Dexvla: Vision-language model with plug-in diffusion expert for general robot control},
  author={Wen, Junjie and Zhu, Yichen and Li, Jinming and Tang, Zhibin and Shen, Chaomin and Feng, Feifei},
  journal={arXiv preprint arXiv:2502.05855},
  year={2025}
}

@article{brohan2022rt,
  title={Rt-1: Robotics transformer for real-world control at scale},
  author={Brohan, Anthony and Brown, Noah and Carbajal, Justice and Chebotar, Yevgen and Dabis, Joseph and Finn, Chelsea and Gopalakrishnan, Keerthana and Hausman, Karol and Herzog, Alex and Hsu, Jasmine and others},
  journal={arXiv preprint arXiv:2212.06817},
  year={2022}
}

@article{liu2023visual,
  title={Visual instruction tuning},
  author={Liu, Haotian and Li, Chunyuan and Wu, Qingyang and Lee, Yong Jae},
  journal={Advances in neural information processing systems},
  volume={36},
  pages={34892--34916},
  year={2023}
}

@article{wen2025tinyvla,
  title={Tinyvla: Towards fast, data-efficient vision-language-action models for robotic manipulation},
  author={Wen, Junjie and Zhu, Yichen and Li, Jinming and Zhu, Minjie and Tang, Zhibin and Wu, Kun and Xu, Zhiyuan and Liu, Ning and Cheng, Ran and Shen, Chaomin and others},
  journal={IEEE Robotics and Automation Letters},
  year={2025},
  publisher={IEEE}
}

@article{ze20243d,
  title={3d diffusion policy: Generalizable visuomotor policy learning via simple 3d representations},
  author={Ze, Yanjie and Zhang, Gu and Zhang, Kangning and Hu, Chenyuan and Wang, Muhan and Xu, Huazhe},
  journal={arXiv preprint arXiv:2403.03954},
  year={2024}
}

@inproceedings{rt2,
  title={Rt-2: Vision-language-action models transfer web knowledge to robotic control},
  author={Zitkovich, Brianna and Yu, Tianhe and Xu, Sichun and Xu, Peng and Xiao, Ted and Xia, Fei and Wu, Jialin and Wohlhart, Paul and Welker, Stefan and Wahid, Ayzaan and others},
  booktitle={Conference on Robot Learning},
  pages={2165--2183},
  year={2023},
  organization={PMLR}
}

@inproceedings{openx2024,
  title={Open x-embodiment: Robotic learning datasets and rt-x models: Open x-embodiment collaboration 0},
  author={O’Neill, Abby and Rehman, Abdul and Maddukuri, Abhiram and Gupta, Abhishek and Padalkar, Abhishek and Lee, Abraham and Pooley, Acorn and Gupta, Agrim and Mandlekar, Ajay and Jain, Ajinkya and others},
  booktitle={2024 IEEE International Conference on Robotics and Automation (ICRA)},
  pages={6892--6903},
  year={2024},
  organization={IEEE}
}

@article{fu2024mobilealoha,
  title={Mobile aloha: Learning bimanual mobile manipulation with low-cost whole-body teleoperation},
  author={Fu, Zipeng and Zhao, Tony Z and Finn, Chelsea},
  journal={arXiv preprint arXiv:2401.02117},
  year={2024}
}

@article{genima,
  title={Generative image as action models},
  author={Shridhar, Mohit and Lo, Yat Long and James, Stephen},
  journal={arXiv preprint arXiv:2407.07875},
  year={2024}
}

@article{maniskill2,
  title={Maniskill2: A unified benchmark for generalizable manipulation skills},
  author={Gu, Jiayuan and Xiang, Fanbo and Li, Xuanlin and Ling, Zhan and Liu, Xiqiang and Mu, Tongzhou and Tang, Yihe and Tao, Stone and Wei, Xinyue and Yao, Yunchao and others},
  journal={arXiv preprint arXiv:2302.04659},
  year={2023}
}

@inproceedings{embodied_robustness,
  title={Robustnav: Towards benchmarking robustness in embodied navigation},
  author={Chattopadhyay, Prithvijit and Hoffman, Judy and Mottaghi, Roozbeh and Kembhavi, Aniruddha},
  booktitle={Proceedings of the IEEE/CVF International Conference on Computer Vision},
  pages={15691--15700},
  year={2021}
}

@article{octo,
  title={Octo: An open-source generalist robot policy},
  author={Team, Octo Model and Ghosh, Dibya and Walke, Homer and Pertsch, Karl and Black, Kevin and Mees, Oier and Dasari, Sudeep and Hejna, Joey and Kreiman, Tobias and Xu, Charles and others},
  journal={arXiv preprint arXiv:2405.12213},
  year={2024}
}

@article{openvla,
  title={Openvla: An open-source vision-language-action model},
  author={Kim, Moo Jin and Pertsch, Karl and Karamcheti, Siddharth and Xiao, Ted and Balakrishna, Ashwin and Nair, Suraj and Rafailov, Rafael and Foster, Ethan and Lam, Grace and Sanketi, Pannag and others},
  journal={arXiv preprint arXiv:2406.09246},
  year={2024}
}

@article{vlaarena,
  title={VLA-Arena: An Open-Source Framework for Benchmarking Vision-Language-Action Models},
  author={Zhang, Borong and Li, Jiahao and Shen, Jiachen and Cai, Yishuai and Zhang, Yuhao and Chen, Yuanpei and Dai, Juntao and Ji, Jiaming and Yang, Yaodong},
  journal={arXiv preprint arXiv:2512.22539},
  year={2025}
}

@inproceedings{robotwin,
  title={Robotwin: Dual-arm robot benchmark with generative digital twins},
  author={Mu, Yao and Chen, Tianxing and Chen, Zanxin and Peng, Shijia and Lan, Zhiqian and Gao, Zeyu and Liang, Zhixuan and Yu, Qiaojun and Zou, Yude and Xu, Mingkun and others},
  booktitle={Proceedings of the computer vision and pattern recognition conference},
  pages={27649--27660},
  year={2025}
}

@misc{chen2025robotwin20scalabledata,
      title={RoboTwin 2.0: A Scalable Data Generator and Benchmark with Strong Domain Randomization for Robust Bimanual Robotic Manipulation}, 
      author={Tianxing Chen and Zanxin Chen and Baijun Chen and Zijian Cai and Yibin Liu and Zixuan Li and Qiwei Liang and Xianliang Lin and Yiheng Ge and Zhenyu Gu and Weiliang Deng and Yubin Guo and Tian Nian and Xuanbing Xie and Qiangyu Chen and Kailun Su and Tianling Xu and Guodong Liu and Mengkang Hu and Huan-ang Gao and Kaixuan Wang and Zhixuan Liang and Yusen Qin and Xiaokang Yang and Ping Luo and Yao Mu},
      year={2025},
      eprint={2506.18088},
      archivePrefix={arXiv},
      primaryClass={cs.RO},
      url={https://arxiv.org/abs/2506.18088}, 
}

@misc{rlbench,
  title={RLBench: The Robot Learning Benchmark and Learning Environment. IEEE Robotics and Automation Letters 5, 2 (2020), 3019--3026},
  author={James, Stephen and Ma, Zicong and Arrojo, David Rovick and Davison, Andrew J},
  year={2020}
}

@article{calvin,
  title={Calvin: A benchmark for language-conditioned policy learning for long-horizon robot manipulation tasks},
  author={Mees, Oier and Hermann, Lukas and Rosete-Beas, Erick and Burgard, Wolfram},
  journal={IEEE Robotics and Automation Letters},
  volume={7},
  number={3},
  pages={7327--7334},
  year={2022},
  publisher={IEEE}
}

@article{simplerenv,
  title={Evaluating real-world robot manipulation policies in simulation},
  author={Li, Xuanlin and Hsu, Kyle and Gu, Jiayuan and Pertsch, Karl and Mees, Oier and Walke, Homer Rich and Fu, Chuyuan and Lunawat, Ishikaa and Sieh, Isabel and Kirmani, Sean and others},
  journal={arXiv preprint arXiv:2405.05941},
  year={2024}
}

@article{safetygym,
  title={Benchmarking safe exploration in deep reinforcement learning},
  author={Ray, Alex and Achiam, Joshua and Amodei, Dario},
  journal={arXiv preprint arXiv:1910.01708},
  volume={7},
  number={1},
  pages={2},
  year={2019}
}

@article{cbf,
  title={Control barrier function based quadratic programs for safety critical systems},
  author={Ames, Aaron D and Xu, Xiangru and Grizzle, Jessy W and Tabuada, Paulo},
  journal={IEEE Transactions on Automatic Control},
  volume={62},
  number={8},
  pages={3861--3876},
  year={2016},
  publisher={IEEE}
}

@book{constrained_mdp,
  title={Constrained Markov decision processes},
  author={Altman, Eitan},
  year={2021},
  publisher={Routledge}
}

@article{saferlsurvey,
  title={A comprehensive survey on safe reinforcement learning},
  author={Garc{\i}a, Javier and Fern{\'a}ndez, Fernando},
  journal={Journal of Machine Learning Research},
  volume={16},
  number={1},
  pages={1437--1480},
  year={2015}
}

@article{adversarial_patch,
  title={Adversarial patch},
  author={Brown, Tom B and Man{\'e}, Dandelion and Roy, Aurko and Abadi, Mart{\'\i}n and Gilmer, Justin},
  journal={arXiv preprint arXiv:1712.09665},
  year={2017}
}

@article{prompt_injection,
  title={Prompt injection attack against llm-integrated applications},
  author={Liu, Yi and Deng, Gelei and Li, Yuekang and Wang, Kailong and Wang, Zihao and Wang, Xiaofeng and Zhang, Tianwei and Liu, Yepang and Wang, Haoyu and Zheng, Yan and others},
  journal={arXiv preprint arXiv:2306.05499},
  year={2023}
}

@article{safebench,
  title={Safebench: A benchmarking platform for safety evaluation of autonomous vehicles},
  author={Xu, Chejian and Ding, Wenhao and Lyu, Weijie and Liu, Zuxin and Wang, Shuai and He, Yihan and Hu, Hanjiang and Zhao, Ding and Li, Bo},
  journal={Advances in Neural Information Processing Systems},
  volume={35},
  pages={25667--25682},
  year={2022}
}

@inproceedings{sapien,
  author    = {Fanbo Xiang and Yuzhe Qin and Kaichun Mo and Yikuan Xia and Hao Zhu and Fangchen Liu and Minghua Liu and Hanxiao Jiang and Yifu Yuan and He Wang and Li Yi and Angel X. Chang and Leonidas J. Guibas and Hao Su},
  title     = {SAPIEN: A SimulAted Part-Based Interactive ENvironment},
  booktitle = {Proceedings of the IEEE/CVF Conference on Computer Vision and Pattern Recognition (CVPR)},
  pages     = {11097--11107},
  year      = {2020},
  month     = jun,
  url       = {https://openaccess.thecvf.com/content_CVPR_2020/html/Xiang_SAPIEN_A_SimulAted_Part-Based_Interactive_ENvironment_CVPR_2020_paper.html}
}

@article{chi2023diffusionpolicy,
  title={Diffusion policy: Visuomotor policy learning via action diffusion},
  author={Chi, Cheng and Xu, Zhenjia and Feng, Siyuan and Cousineau, Eric and Du, Yilun and Burchfiel, Benjamin and Tedrake, Russ and Song, Shuran},
  journal={The International Journal of Robotics Research},
  volume={44},
  number={10-11},
  pages={1684--1704},
  year={2025},
  publisher={Sage Publications Sage UK: London, England}
}

@article{zhao2023act,
  title={Learning fine-grained bimanual manipulation with low-cost hardware},
  author={Zhao, Tony Z and Kumar, Vikash and Levine, Sergey and Finn, Chelsea},
  journal={arXiv preprint arXiv:2304.13705},
  year={2023}
}

@article{liu2024rdt,
  title={Rdt-1b: a diffusion foundation model for bimanual manipulation},
  author={Liu, Songming and Wu, Lingxuan and Li, Bangguo and Tan, Hengkai and Chen, Huayu and Wang, Zhengyi and Xu, Ke and Su, Hang and Zhu, Jun},
  journal={arXiv preprint arXiv:2410.07864},
  year={2024}
}

@article{physicalintelligence2024pi0,
  title={$\pi$0: A Vision-Language-Action Flow Model for General Robot Control},
  author={Black, Kevin and Brown, Noah and Driess, Danny and Esmail, Adnan and Equi, Michael and Finn, Chelsea and Fusai, Niccolo and Groom, Lachy and Hausman, Karol and Ichter, Brian and others},
  journal={arXiv preprint arXiv:2410.24164},
  year={2024}
}

@article{wang2025safety,
  title={Safety of embodied navigation: A survey},
  author={Wang, Zixia and Hu, Jia and Mu, Ronghui},
  journal={arXiv preprint arXiv:2508.05855},
  year={2025}
}

@article{sermanet2025generating,
  title={Generating robot constitutions \& benchmarks for semantic safety},
  author={Sermanet, Pierre and Majumdar, Anirudha and Irpan, Alex and Kalashnikov, Dmitry and Sindhwani, Vikas},
  journal={arXiv preprint arXiv:2503.08663},
  year={2025}
}

@article{gruner2025towards,
  title={Towards Safe Robot Foundation Models},
  author={Gruner, Theo and Palenicek, Daniel and Liu, Puze and Watson, Joe and Tateo, Davide and Peters, Jan and others},
  journal={arXiv e-prints},
  pages={arXiv--2503},
  year={2025}
}

@article{wei2024ensuring,
  title={Ensuring force safety in vision-guided robotic manipulation via implicit tactile calibration},
  author={Wei, Lai and Ma, Jiahua and Hu, Yibo and Zhang, Ruimao},
  journal={arXiv preprint arXiv:2412.10349},
  year={2024}
}

@article{intelligence2025pi0,
  title={$\pi$0. 5: A vision-language-action model with open-world generalization. arXiv 2025},
  author={Intelligence, P and Black, K and Brown, N and Darpinian, J and Dhabalia, K and Driess, D and Esmail, A and Equi, M and Finn, C and Fusai, N and others},
  journal={arXiv preprint arXiv:2504.16054},
  year={2025}
}

@article{lasota2017survey,
  title={A survey of methods for safe human-robot interaction},
  author={Lasota, Przemyslaw A and Fong, Terrence and Shah, Julie A},
  journal={Foundations and Trends{\textregistered} in Robotics},
  volume={5},
  number={4},
  pages={261--349},
  year={2017},
  publisher={Emerald Publishing Limited}
}

@article{brunke2022safe,
  title={Safe learning in robotics: From learning-based control to safe reinforcement learning},
  author={Brunke, Lukas and Greeff, Melissa and Hall, Adam W and Yuan, Zhaocong and Zhou, Siqi and Panerati, Jacopo and Schoellig, Angela P},
  journal={Annual Review of Control, Robotics, and Autonomous Systems},
  volume={5},
  number={1},
  pages={411--444},
  year={2022},
  publisher={Annual Reviews}
}

@article{ji2023safety,
  title={Safety gymnasium: A unified safe reinforcement learning benchmark},
  author={Ji, Jiaming and Zhang, Borong and Zhou, Jiayi and Pan, Xuehai and Huang, Weidong and Sun, Ruiyang and Geng, Yiran and Zhong, Yifan and Dai, Josef and Yang, Yaodong},
  journal={Advances in Neural Information Processing Systems},
  volume={36},
  pages={18964--18993},
  year={2023}
}

@article{yuan2022safe,
  title={Safe-control-gym: A unified benchmark suite for safe learning-based control and reinforcement learning in robotics},
  author={Yuan, Zhaocong and Hall, Adam W and Zhou, Siqi and Brunke, Lukas and Greeff, Melissa and Panerati, Jacopo and Schoellig, Angela P},
  journal={IEEE Robotics and Automation Letters},
  volume={7},
  number={4},
  pages={11142--11149},
  year={2022},
  publisher={IEEE}
}

@article{amodei2016concrete,
  title={Concrete problems in AI safety},
  author={Amodei, Dario and Olah, Chris and Steinhardt, Jacob and Christiano, Paul and Schulman, John and Man{\'e}, Dan},
  journal={arXiv preprint arXiv:1606.06565},
  year={2016}
}

@article{liu2023libero,
  title={Libero: Benchmarking knowledge transfer for lifelong robot learning},
  author={Liu, Bo and Zhu, Yifeng and Gao, Chongkai and Feng, Yihao and Liu, Qiang and Zhu, Yuke and Stone, Peter},
  journal={Advances in Neural Information Processing Systems},
  volume={36},
  pages={44776--44791},
  year={2023}
}

\end{document}